\documentclass{article}
\usepackage{amsmath,amssymb}

\usepackage{textcomp}
\usepackage{dsfont}
\usepackage{subcaption}
\usepackage{bm}
\usepackage{graphicx}
\usepackage{fourier}
\usepackage{textcomp}
\graphicspath{{./figs/}}
\usepackage{natbib}
\usepackage[hyphens]{url}   
\usepackage{hyperref}       
\usepackage{booktabs}       
\usepackage[final]{neurips_2021}

\title{Uncertainty estimation for out-of-distribution detection in computational histopathology}

\author{
  Lea Goetz \\
  Artificial Intelligence and Machine Learning, GSK, London \\
  \texttt{lea.x.goetz@gsk.com} \\
}

\begin{document}
\maketitle

\begin{abstract}
  In computational histopathology algorithms now outperform humans on 
  a range of tasks, but to date none are
  employed for automated diagnoses in the clinic. 
  Before algorithms can be involved in such high-stakes decisions 
  they need to "know when they don't know", i.e., they need to estimate their
  predictive uncertainty. This allows them to defer potentially erroneous
  predictions to a human pathologist, thus increasing their safety.
  Here, we evaluate the predictive performance and calibration of several
  uncertainty estimation methods on clinical histopathology data. 
  We show that a distance-aware uncertainty estimation method outperforms commonly
  used approaches, such as Monte Carlo dropout and deep ensembles.
  However, we observe a drop in predictive performance and calibration on novel samples
  across all uncertainty estimation methods tested.
  We also investigate the use of uncertainty thresholding to reject out-of-distribution 
  samples for selective prediction. We demonstrate the limitations of this approach 
  and suggest areas for future research.
\end{abstract}

\section{Introduction}
Over the last decade, computational histopathology has seen a surge in algorithms achieving
equal or superior performance compared to human pathologists on a diverse set of tasks, such as 
metastasis detection \citep{liu2019artificial}, prediction of molecular markers
from tissues \citep{naik2020deep}, 
and patient survival \citep{mobadersany2018predicting}. However,
despite these academic successes, none of these models have to date been used in a 
decision making capacity in a clinical setting, and only one software is approved for 
assisting pathologists\footnote{\url{https://www.fda.gov/news-events/press-announcements/fda-authorizes-software-can-help-identify-prostate-cancer}}.
What explains this \textit{translation gap}?
In addition to model-independent hurdles (e.g.,
regulatory approval, integration into clinical workflows, etc. \citep{steiner2021closing}), 
a key requirement for models in high-stakes applications is 
robustness to data shift.
However, in histopathology datasets are generally small (compared to standard ML datasets, such as ImageNet)
and models trained on them are 
more likely to overfit low-level features 
\citep{Arpit_memorization}, such as texture \citep{Geirhos2019_texturebias},
which do not generalize to novel datasets.
At the same time, large distribution shifts between training and test datasets 
are common in histopathology \citep{zech2018variable,albadawy2018deep} 
as a result of tissue preprocessing and image acquisition 
\citep{veta2016,komura2018machine,tellez2019quantifying}. 

As building robustness against these shifts into models is difficult, they need
to \textit{"know when they don't know"}
\citep{shafaei2018less,roy2022does}, i.e., to estimate the uncertainty
of their predictions. 
We review relevant use cases 
and how methods for uncertainty estimation have been applied in computational histopathology 
to date.

\subsection{Related work}
There are (at least) two scenarios where a model has a high predictive 
uncertainty: when it encounters an unknown sample (out-of-distribution, OOD) 
 or a known (in-distribution, ID) but ambiguous sample. 
 In either case, the model accuracy is likely low and samples should be 
deferred to a human expert 
 to avoid erroneous predictions. 
By setting an upper threshold on the uncertainty of samples, uncertainty estimation 
methods can be used for selective prediction. 

\paragraph*{Uncertainty estimation methods}
The maximum softmax probability (MSP)
\citep{hendrycks2016baseline} is a common baseline estimate of predictive uncertainty,
but in general not well calibrated \citep{tempscaling}. 
Bayesian neural networks \citep{Mackay1992APB,Buntine1991BayesianB,Mackay1995ProbableNA}
provide a principled approach to quantify model uncertainty  
but require dedicated architectures, are difficult and expensive to train,
hard to scale to large models and datasets, and their uncertainty estimates may not be robust to to dataset shift 
\citep{Ovadia2019CanYT,gustafsson2020evaluating}.
There are various approximations that reduce the computational complexity, 
such as low rank approximations \citep{dusenberry20a_efficient_BNNs} 
and Markov chain Monte Carlo methods \citep{Welling2011BayesianLV},
or that can be transferred to standard network architectures, 
such as Laplace approximations \citep{mackay1992bayesian,laplace2021}.

Monte-Carlo (MC) dropout \citep{gal2016dropout} is a widely used method, as it is 
easily implemented in architectures with DropOut layers \citep{hinton2012improving}. 
Currently, state-of-the-art uncertainty estimates are obtained by using the entropy in the predictions of 
a deep ensemble \citep{deepensembles}, or an efficient approximation thereof \citep{Wen2020BatchEnsembleAA}.
Similarly competitive are a number of recently proposed methods that require 
only a single-forward pass and are "distance-aware" \citep{tagasovska2019single,sngp,ddu,duq,jain2021deup}.
They use feature space distances between training and test samples to 
quantify uncertainty. This allows them to accurately estimate uncertainty far away from the decision boundary.

\paragraph*{Uncertainty estimation and OOD detection in histopathology}
Several studies in histopathology use deep ensembles \citep{deepensembles, pocevivciute2021can, thagaard2020can},
while \cite{senousy20213e} use MSP to select models for an ensemble. 
\cite{Linmans2020EfficientOD} use an ensembling approach on multiple prediction heads
for open set recognition (OSR).
\cite{rkaczkowski2019ara} show that MC Dropout-based uncertainty is high for
ambiguous or mislabelled patches, but did not test on OOD data; 
\cite{syrykh2020accurate} use MC Dropout for both OSR  
and OOD detection but do not report OOD detection metrics.
Note that, MC Dropout can be problematic in network architectures commonly used in computational histopathology 
\footnote{\label{footnote_batchnorm_dropout} \cite{MCdropout_vs_BatchNorm} demonstrated
that applying MC Dropout with dropout rates \begin{math}
  \geq 0.1
\end{math} in networks that use Layer Normalization \citep{layernorm} --
as is the case in the related work cited above --
can be problematic: the combination causes unstable numerical behavior during inference
on a range of architectures (DenseNet, ResNet, ResNeXt \citep{resnext} and Wide ResNet \citep{wide_resnet}), 
and requires additional implementation strategies}, 
and has been shown to negatively affect task performance \citep{Linmans2020EfficientOD}. 

Unfortunately, the OOD detection reported in \cite{syrykh2020accurate, pocevivciute2021can, senousy2021mcua} is of limited insight, 
as uncertainty thresholds for selective prediction were set on the same OOD data on which 
performance was evaluated. 
\cite{dolezal2022uncertainty} avoid such data leakage by setting the uncertainty 
threshold on validation data using cross-validation. However, it is unclear whether a 
threshold chosen to distinghish between correct and incorrect ID samples 
is suitable to separate ID and OOD data.
For example, on a dataset for which there is no correct diagnosis, 
still more than 20\% of slides are rated high-confidence by the uncertainty-aware classifier of \cite{dolezal2022uncertainty}.

\subsection{Contributions}
The uncertainty estimation methods used by previous work like MC Dropout
or deep ensembles don't show state-of-the-art performance on standard ML datasets
or require substantial additional compute, respectively.
They also estimate uncertainty around the decision boundary, i.e. are most suitable 
to detect ambiguous samples, but may give high confidence estimates for OOD samples far away from the decision boundary.
Recently proposed "distance-aware" uncertainty estimation methods 
\citep{tagasovska2019single,sngp,ddu,duq} address these concerns and
present an attractive alternative for histopathology.
To the best of our knowledge they have not yet been evaluated on 
challenging clinical datasets. Because of its superior performance and 
relative ease of implementation in combination with existing architectures, we chose 
a spectral-normalized Gaussian Process (SNGP) \citep{sngp} as a representative distance-aware method
and compare its performance to methods currently widely used in histopathology.

Our paper makes the following contributions:
\begin{itemize}
  \item We evaluate the predictive performance and calibration of a baseline (MSP),
        two commonly used (MC Dropout and deep ensembles),
        and one distance-aware uncertainty estimation method (SNGP) on CIFAR-10 and 
        compare to two datasets of clinical histopathology data.
  \item We demonstrate the limitations of using uncertainty thresholding 
        for OOD detection in histopathology; we discuss caveats and areas for further research 
        in applying uncertainty estimation in histopathology.
\end{itemize}

\section{Methods}
\subsection{Datasets}

First, we evaluate the uncertainty estimation methods on CIFAR-10, to 
investigate whether their performance on a standard ML dataset transfers to clinical histopathology data.
As CIFAR-10 OOD data, we designed image corruptions that emulate realistic 
histopathological distribution shifts. Using Pytorch's ColorJitter transform,
we randomly change the brightness, contrast, saturation and hue
based on ranges we observed in clinical whole-slide images (WSIs) 
stained with hematoxin and eosin (H\&E) dye 
(brightness=0, contrast=0, saturation=0.1, hue=0.1). 

 \begin{figure}[h]
   \includegraphics[angle=0,width=\textwidth]{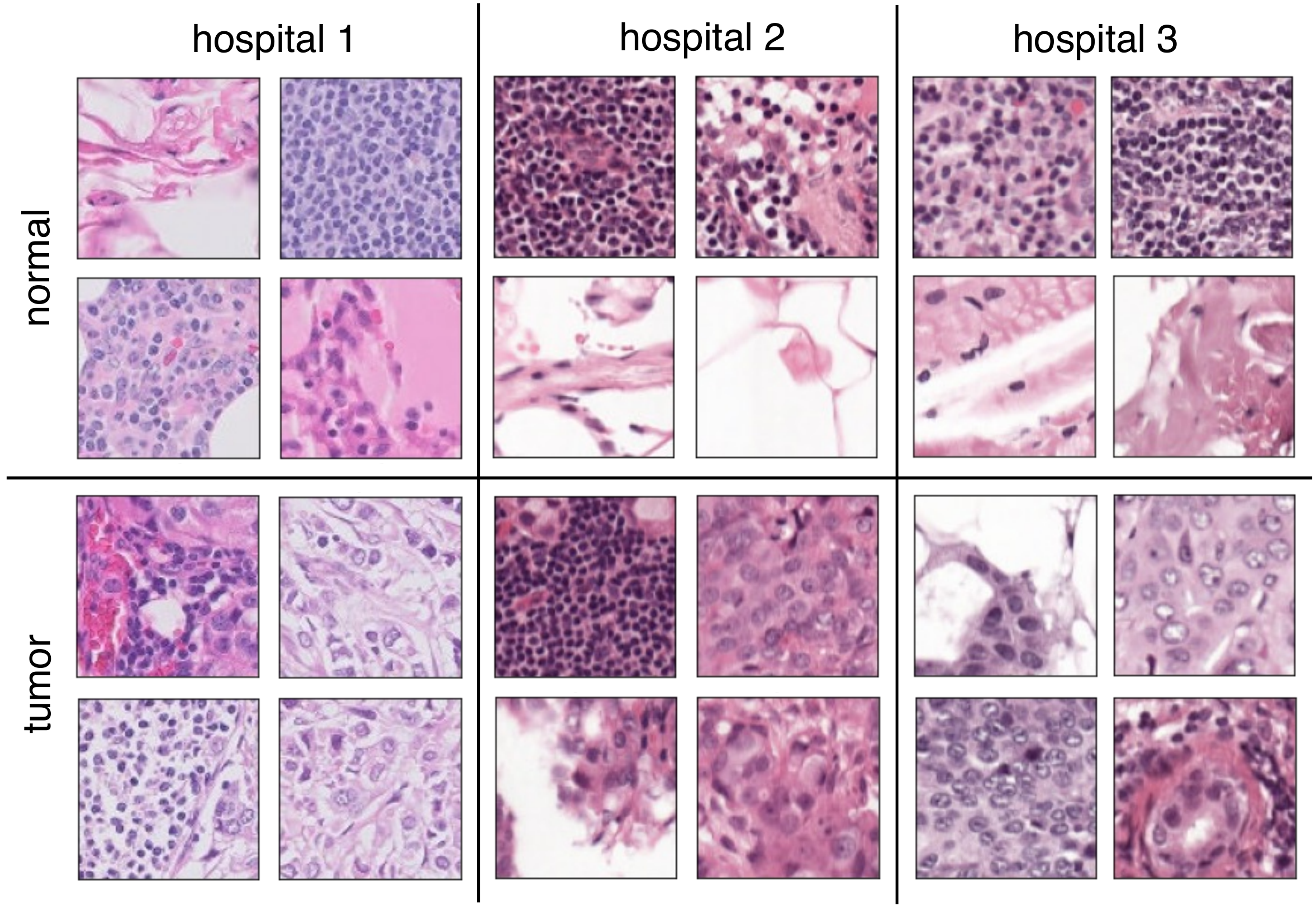}
   \caption{Representative tiles of WSIs from the Camelyon17 dataset, 
   illustrating the variation between normal and tumour tissue, and between different hospitals
   used as training/validation datasets. While
   from the same hospitals, training and validation sets are non-overlapping.
   } \label{figcamelyon17overview}
 \end{figure}

 \begin{figure}[h]
    \includegraphics[angle=0,width=\textwidth]{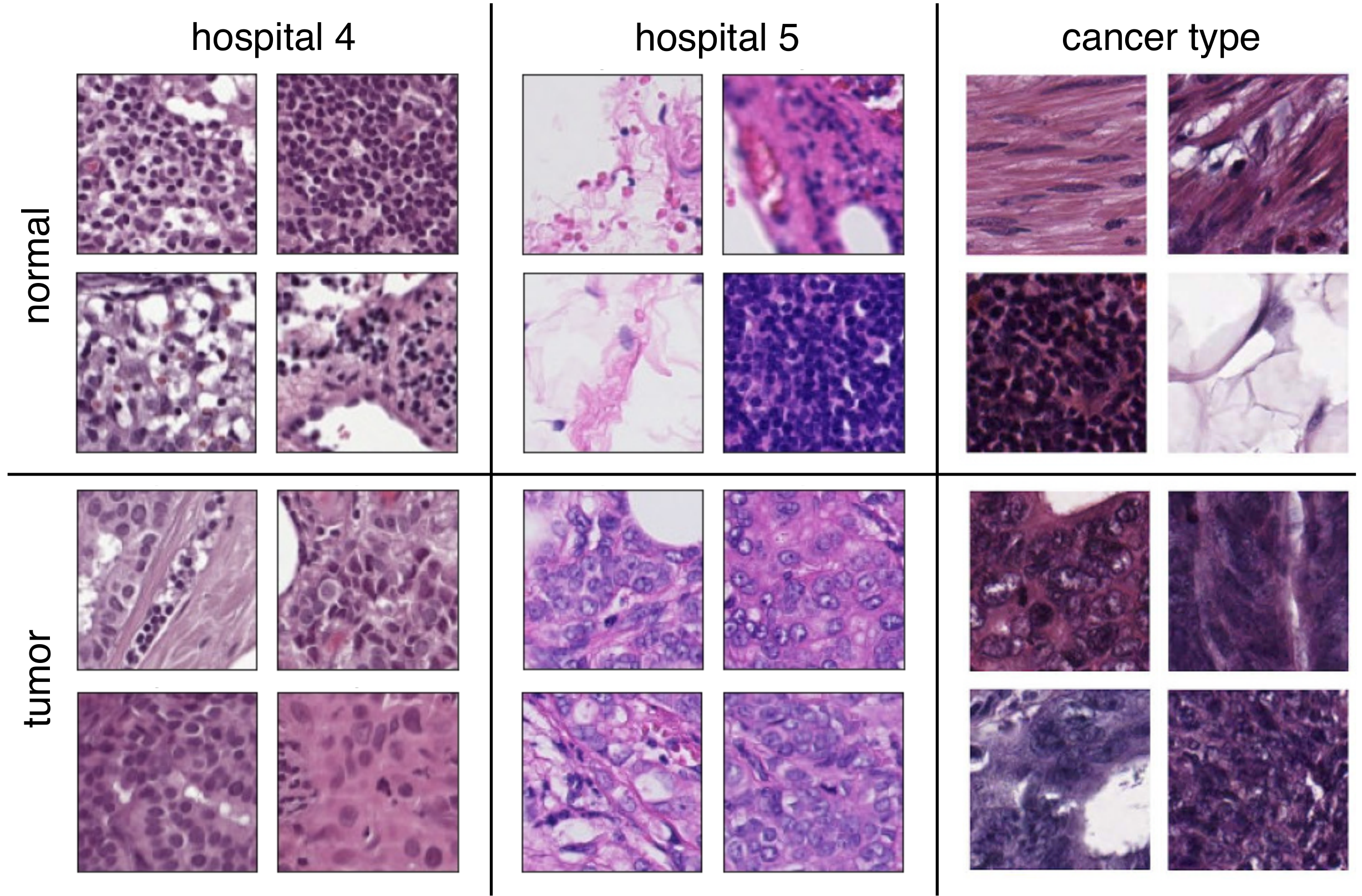}
    \caption{Sample tiles from the Camelyon17 and colorectal cancer dataset \citep{kather2016collection}, 
    used as OOD datasets.
    } \label{figcamelyon17_2_overview}
  \end{figure}

Second, we use a patch-based version \citep{bandi2018detection} of the Camelyon17 
grand challenge dataset (\url{https://camelyon17.grand-challenge.org/}), which contains patches of 
WSIs of H\&E stained lymph node sections.
This dataset is open source and available under a CC0 1.0 license\footnote{\url{https://creativecommons.org/publicdomain/zero/1.0}}.
Following \citep{wilds2021}, we divide the dataset into four folds according 
to the hospitals where WSIs were obtained:
the training set consists of WSIs taken from three hospitals (hospitals 1--3), 
each contributing ten WSIs for a total of 302 436 patches.
We use three OOD test sets: OOD1 (hospital 4), consisting of 34 904 patches taken from 10 WSIs of a 
different hospital; 
OOD2 (hospital 5), consisting of 85 054 patches taken from 10 WSIs of another different 
hospital\footnote{Note that in previous work \citep{wilds2021}, OOD1 and OOD2 datasets have been considered
"near" and "far" OOD, respectively, based on the more different visual appearance of patches from hospital 5.}. 
Finally, we evaluate performance to detect a different tumour type on colorectal cancer data from \cite{kather2016collection},
which is available under a CC BY 4.0 license\footnote{\url{https://creativecommons.org/licenses/by/4.0/legalcode}}.
For both the Camelyon17 and colorectal cancer datasets the human biological samples were sourced ethically and
their research use was in accord with the terms of the informed
consents under an IRB/EC approved protocol.

\subsection{Uncertainty estimation methods}
As a baseline we use MSP \citep{hendrycks2016baseline}, because it is model-agnostic, 
and requires no additional implementation effort as it can be read-out from final layer logits.
To compare to prior work in histopathology, we use deep ensembles \citep{deepensembles}
and MC Dropout \citep{gal2016dropout} following implementations
in \cite{thagaard2020can} and \cite{BatchBALD}. For MC Dropout, 
we add a DropOut layer (p=0.5, \citep{hinton2012improving}) 
before the final layer, and perform several forward passes to generate a distribution over predictions
(we use 32 MC samples \citep{Ovadia2019CanYT}). We use ensembles with four members.
For a brief comparison of uncertainty estimation method characteristics see Table \ref{ue-methods-table}. 

We also implemented   
a single forward pass, distance-aware method, SNGP \citep{sngp}. This required modifications to the model architecture: 
we added a Gaussian Process layer
with random feature approximation adapted from \url{https://github.com/google/edward2}, 
following \cite{rahimi_recht_random_features}. We also added spectral normalization \citep{spectral_normalization}
to the training.
For all models, we train a ResNet-50 \citep{resnet} for 100 epochs 
to predict tumor vs normal, using a cross-entropy loss
with Adam \citep{kingma2014adam} (lr=1e-3, weight decay=1e-5) as an optimizer.

\begin{table}
  \centering
  \begin{tabular}{llllll}
    \toprule
    \textbf{UE method}  & \textbf{model-agnostic}  & \textbf{single pass}  & \textbf{distance-aware} & \textbf{implementation}  \\
    \cmidrule(r){1-5}
    Softmax           &             yes                     &	yes               &	 no       & logits     \\
    Dropout        &               yes*               & no     &	no  & DropOut Layer; $n$ passes \\
    Ensemble    & yes  & no  & no & $n$ random initializations \\
    SNGP      & no & yes & yes & GP Layer; train with SN \\    \bottomrule
  \end{tabular}
  \caption{\label{ue-methods-table} Comparison of uncertainty estimation methods evaluated. 
  DropOut Layer: \url{https://pytorch.org/docs/stable/generated/torch.nn.Dropout.html}; 
  GP Layer: \url{https://github.com/google/edward2};
  SN = spectral normalization;
  * Potential architectural incompatibilities, see \citep{MCdropout_vs_BatchNorm}.}
\end{table}

\subsection{Evaluation metrics}
For a dataset of $N$ input-output pairs $(x_n, y_n)$, 
in our case with a binary label $y_n$, 
we evaluate predicitive accuracy, average precision (AP)
\footnote{
Average precision is less sensitive to class imbalance than accuracy.
While our ID validation data, OOD1 and OOD2 datasets are balanced,
the colorectal cancer dataset has a 7:1 ratio of normal:tumor tissue.}
, Expected Calibration Error (ECE) 
and Maximum Calibration Error (MCE) \citep{naeini2015obtaining}, 
which are defined as follows:

\begin{equation}
  \text{AP} = \sum_{k=1}^K \text{P}(k)\Delta r(k)
\end{equation}

where $P(k)$ is the is the precision at threshold $k$ and 
$\Delta r(k)$ is the change in recall from $k-1$ to $k$,

\begin{equation}
  \text{ECE} = \sum_{b=1}^B \frac{n_b}{N} | \text{acc}(b) - \text{con}(b) |
\end{equation}

\begin{equation}
  \text{MCE} = \max_{b=1,...,B} \frac{n_b}{N} | \text{acc}(b) - \text{con}(b) |
\end{equation}

where $\text{acc}(b)$ and $\text{con}(b)$ 
are the accuracy and confidence of bin $b$ respectively,
$n_b$ number is the number of predictions in bin $b$, and $B=15$ is the number of bins.

We also use the AUROC score of distinguishing between ID and OOD data 
\citep{deepensembles,ddu}.  
All of the above metrics are computed on classifier logits. 
We report results as mean $\pm$ std, averaged over four random initializations. For ensembles, 
we report results averaged over three ensembles with four members each, i.e., 12 random initializations in total.

\subsection{Uncertainty thresholding}
To set an uncertainty threshold for selective prediction, we chose Youden's J statistic \citep{Youden1950IndexFR}
because it optimally trades of sensitivity and specificity, and is widely used in diagnostic tests (e.g. \citep{perkins2005youden}). 
Youden's J is defined as follows:

\begin{equation}
  \text{J} = \frac{\text{TP}}{\text{TP} + \text{FN}} + \frac{\text{TN}}{\text{TN} + \text{FP}} - 1
\end{equation}

where TP are true positives, FN are false negatives, TN are true negatives and FP are false positives.

\begin{table}
  \resizebox{\textwidth}{!}{
  \centering
    \begin{tabular}{|l|l|l|l|l|l|l|}
    \hline
    \textbf{UE method}     & \textbf{Accuracy \textuparrow}    & \textbf{AP \textuparrow}  & \textbf{ECE \textdownarrow}  & \textbf{MCE \textdownarrow} & \textbf{AUROC-ood \textuparrow} \\
    \hline
    \multicolumn{6}{|c|}{validation}                   \\
    \hline
    Softmax   &  0.870 $\pm$ 0.004 & 0.918 $\pm$ 0.003 & 0.089 $\pm$ 0.003 & 0.433 $\pm$ 0.192 & - \\ 
    Dropout   &  0.868 $\pm$ 0.004 & 0.919 $\pm$ 0.003 & 0.089 $\pm$ 0.003 & \textbf{0.319 $\pm$ 0.017} & - \\ 
    Ensemble  &  \textbf{0.912 $\pm$ 0.002} & \textbf{0.942 $\pm$ 0.005} & \textbf{0.049 $\pm$ 0.004} & \textbf{0.257 $\pm$ 0.050} & - \\ 
    SNGP      &  0.854 $\pm$ 0.008 & 0.906 $\pm$ 0.007 & 0.104 $\pm$ 0.006 & 0.451 $\pm$ 0.171 & - \\ 
    \hline
    \multicolumn{6}{|c|}{OOD}                   \\
    \hline
    Softmax   &  \textbf{0.490 $\pm$ 0.020} & 0.618 $\pm$ 0.008 & \textbf{0.370 $\pm$ 0.023} & 0.520 $\pm$ 0.017 & 0.785 $\pm$ 0.007  \\
    Dropout   &  0.426 $\pm$ 0.125 & 0.548 $\pm$ 0.163 & 0.449 $\pm$ 0.166 & 0.576 $\pm$ 0.125 & 0.783 $\pm$ 0.024   \\
    Ensemble  &  \textbf{0.519 $\pm$ 0.013} & \textbf{0.694 $\pm$ 0.003} & \textbf{0.328 $\pm$ 0.026} & 0.508 $\pm$ 0.094 & \textbf{0.819 $\pm$ 0.002}  \\
    SNGP      &  0.455 $\pm$ 0.024 & 0.562 $\pm$ 0.020 & 0.409 $\pm$ 0.032 & 0.545 $\pm$ 0.033 & 0.755 $\pm$ 0.015 	 \\
    \hline
  \end{tabular}}
  \caption{\label{cifar-table} Results on CIFAR-10 validation and CIFAR-10-OOD dataset (= histopathology-like shift), mean $\pm$ std,  averaged over 4 seeds.
   Highest performing method (> 1 std) in \textbf{bold}.}
\end{table}

\section{Experiments}
\subsection{Predictive performance and calibration}
Deep ensembles outperform all other methods both in terms of 
predictive performance and calibration on both the CIFAR-10 ID and OOD set, 
however, the drop in performance on OOD data is substantial. 
The results on CIFAR-10 
(Table \ref{cifar-table}) 
only somewhat generalize:
on histopathology data all methods show strong performance on ID samples, but predictive performance 
and in particular calibration drops sharply 
for the far OOD datasets (hospital 5 and cancer type). 
Different uncertainty estimation methods 
perform best on different datasets and metrics. Notably, SNGP outperforms all other methods 
on the hospital 5 OOD dataset. 

\begin{table}
  \begin{tabular}{|l|l|l|l|l|l|l|}
    \hline
    \textbf{UE method}     & \textbf{Accuracy \textuparrow}    & \textbf{AP \textuparrow}  & \textbf{ECE \textdownarrow}  & \textbf{MCE \textdownarrow} & \textbf{AUROC-ood \textuparrow} \\
    \hline
    \multicolumn{6}{|c|}{validation}                   \\
    \hline
    Softmax     & 0.984 $\pm$ 0.015 & \textbf{0.998 $\pm$ 0.003} & 0.009 $\pm$ 0.010 & 0.192 $\pm$ 0.038 & - \\ 
    Dropout     & 0.990 $\pm$ 0.001 & 0.986 $\pm$ 0.002 & 0.005 $\pm$ 0.001 & 0.154 $\pm$ 0.036 & - \\ 
    Ensemble    & \textbf{0.995 $\pm$ 0.000} & \textbf{1.000 $\pm$ 0.000} & 0.002 $\pm$ 0.000 & 0.190 $\pm$ 0.056 & - \\       
    SNGP        & 0.990 $\pm$ 0.001 & 0.999 $\pm$ 0.000 & \textbf{0.001 $\pm$ 0.000} & \textbf{0.069 $\pm$ 0.010} & - \\ 
    \hline
    \multicolumn{6}{|c|}{OOD 1 (hospital 4)}                   \\
    \hline
    Softmax     & 0.765 $\pm$ 0.079 & 0.793 $\pm$ 0.082 & 0.211 $\pm$ 0.074 & 0.335 $\pm$ 0.041 & 0.623 $\pm$ 0.057 \\
    Dropout     & \textbf{0.801 $\pm$ 0.052} & 0.769 $\pm$ 0.040 & 0.177 $\pm$ 0.051 & 0.340 $\pm$ 0.031 & 0.637 $\pm$ 0.083 \\
    Ensemble    & \textbf{0.832 $\pm$ 0.019} & \textbf{0.876 $\pm$ 0.032} & 0.148 $\pm$ 0.019 & 0.320 $\pm$ 0.017 & 0.632 $\pm$ 0.021 \\
    SNGP        & 0.793 $\pm$ 0.065 & \textbf{0.909 $\pm$ 0.031} & 0.125 $\pm$ 0.063 & \textbf{0.215 $\pm$ 0.075} & \textbf{0.830 $\pm$ 0.042} \\
    \hline
    \multicolumn{6}{|c|}{OOD 2 (hospital 5)}                   \\
    \hline
    Softmax     & 0.624 $\pm$ 0.054 & 0.662 $\pm$ 0.072 & 0.325 $\pm$ 0.061 & 0.363 $\pm$ 0.065 & 0.765 $\pm$ 0.015 \\
    Dropout     & \textbf{0.633 $\pm$ 0.074} & 0.597 $\pm$ 0.057 & 0.322 $\pm$ 0.072 & 0.396 $\pm$ 0.077 & 0.776 $\pm$ 0.042 \\
    Ensemble    &  0.619 $\pm$ 0.020 & 0.665 $\pm$ 0.011 & 0.341 $\pm$ 0.024 & 0.385 $\pm$ 0.008 & 0.775 $\pm$ 0.042 \\
    SNGP       & \textbf{0.737 $\pm$ 0.038} & \textbf{0.809 $\pm$ 0.045} & \textbf{0.156 $\pm$ 0.032} & \textbf{0.229 $\pm$ 0.031} & \textbf{0.887 $\pm$ 0.032} \\
    \hline
    \multicolumn{6}{|c|}{OOD 3 (cancer type)}                   \\
    \hline
    Softmax     & 0.591 $\pm$ 0.130 & 0.120 $\pm$ 0.006 & 0.353 $\pm$ 0.109 & 0.584 $\pm$ 0.072 & 0.718 $\pm$ 0.046 \\
    Dropout     & 0.675 $\pm$ 0.102 & 0.123 $\pm$ 0.002 & 0.286 $\pm$ 0.085 & \textbf{0.334 $\pm$ 0.064} & \textbf{0.760 $\pm$ 0.081} \\
    Ensemble    & 0.637 $\pm$ 0.021 & 0.124 $\pm$ 0.000 & 0.320 $\pm$ 0.013 & \textbf{0.379 $\pm$ 0.033} & 0.766 $\pm$ 0.031 \\
    SNGP       & 0.631 $\pm$ 0.057 & 0.124 $\pm$ 0.001 & 0.304 $\pm$ 0.060 & 0.432 $\pm$ 0.089 & \textbf{0.851 $\pm$ 0.019} \\
    \hline
  \end{tabular}
  \caption{Results on Camelyon17 and \citep{kather2016collection} datasets, mean $\pm$ std averaged over 4 seeds. 
  Highest performing method (> 1 std) in \textbf{bold}.
  }\label{camelyontable}
\end{table}

\subsection{OOD detection}
All tested UE methods perform poorly on OOD data, in particular the more
different the OOD samples are from the 
training set, even though this "far OOD" detection is considered an easier task 
than "near OOD" detection \citep{shafaei2018less}. 
SNGP has similar performance across the different OOD datasets
and consistently outperforms all other methods in OOD detection.

What causes this superior performance? 
SNGP makes use of feature space distances between training and test samples to 
quantify uncertainty: it combines a distance-preserving feature extractor, 
(spectral normalization \citep{spectral_normalization})
with a distance-aware classifier (Gaussian Process \citep{sngp}),
i.e., by design it should perform well on distinguishing between OOD and ID data. 
In addition, our results suggest that the uncertainty estimates of SNGP are also well calibrated on ID data.

\subsection{Uncertainty thresholding}
\begin{figure}[h]
    \includegraphics[angle=-90,width=\textwidth]{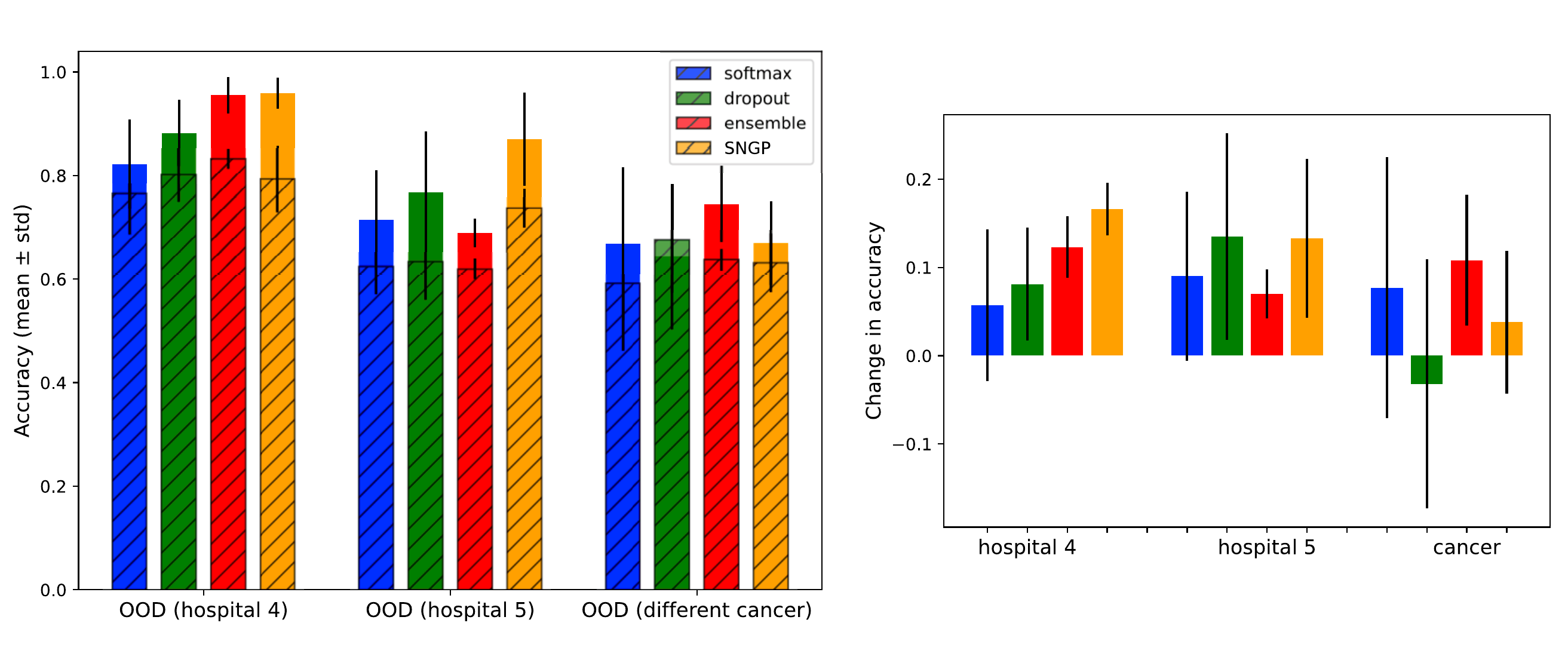}
    \caption{Left: accuracy before (hashed) and after (unhashed bars) uncertainty thresholding. Right:
    change in accuracy due to thresholding. For results on average precision, see Appendix.
    } \label{figaccbars}
  \end{figure}

For selective prediction, we set an uncertainty threshold to separate 
OOD and ID data. We chose not to set this threshold on ID data;
however, setting it on OOD data because it is unclear that this separates well between ID and OOD samples.
Instead, we evaluate how well OOD thresholds set on different OOD datasets separate ID from OOD data. 
In Figure \ref*{figaccbars} we report results for
OOD detection on hospital 4 with threshold set on hospital 5; as well as OOD detection on hospital 5 and 
a different cancer type, with threshold set on hospital 4. 
All other combinations can be found in the Appendix.
While uncertainty thresholding does in general increase accuracy, 
the gains are very variable across datasets and methods. 
Notably a large proportion of samples are rejected across all methods and thresholds (see Table \ref{fraction-retained-table}),
putting into question the practicality of using uncertainty thresholding in clinical practice.

\begin{table}
  \centering
    \begin{tabular}{|l|l|l|l|l|}
    \hline
    \textbf{UE method}     & \textbf{hospital 4}    & \textbf{hospital 5}  & \textbf{cancer type} \\
    \hline
    Softmax   &  0.658 $\pm$ 0.110  & 0.234 $\pm$ 0.215 & 0.278 $\pm$ 0.208 \\
    Dropout   &   0.490 $\pm$ 0.110 & 0.158 $\pm$ 0.133 & 0.211 $\pm$ 0.130 \\
    Ensemble  &  0.357 $\pm$ 0.209 & 0.355 $\pm$ 0.237 & 0.344 $\pm$ 0.205 \\
    SNGP      & 0.341 $\pm$ 0.114 & 0.169 $\pm$ 0.049 & 0.292 $\pm$ 0.134 \\
    \hline
  \end{tabular}
  \caption{\label{fraction-retained-table} Fraction of samples (mean $\pm$ std) retained after rejection based on 
  an uncertainty threshold that optimizeds Youden's J, set on hospital 5 and hospital 4 
  for hospital 4 and hospital 5 \& cancer type, respectively.}
\end{table}

\section{Discussion}
\subsection{CIFAR-10 vs Camelyon-17/colorectal cancer datasets}
We found that the performance of uncertainty estimation methods observed on CIFAR-10 
does not generalize well to histopathology datasets. 
This is in line with previous work demonstrating that
several methods which improve model performance and robustness on standard ML datasets
have no effect, or decrease model performance, on histopathology data \citep{Tamkin2022ActiveLH}.
Furthermore, model performance on clinical histopathology datasets 
exhibits high variance across seeds \citep{wilds2021}, 
likely also as a result of the highly variable sample collection and processing. 

Common ML datasets are very different from clinical histopathology data: 
first, the CIFAR-10 dataset contains images of a single, centered object, 
at a low resolution, whereas tiles of histopathology slides often contain many different objects 
and textures at a high resolution. 
Second, while there is some noise associated with CIFAR-10 labels 
\citep*{peterson2019human,wei2021learning}, 
the variability of individual pathologists and across pathologists 
in classifying histopathology images is much larger \citep{Gomes2014InterobserverVB,TanReproducibility}.
Thus, evaluating models that perform well on standard ML datasets on a range of histopathological data
is an important research direction for bringing state-of-the-art ML performance closer 
to having impact in the clinical practice. 

\subsection{Uncertainty thresholding}
While we and others have demonstrated that setting an uncertainty threshold can be used 
to increase predictive accuracy, both on ID and OOD data, it remains an open problem how 
to set this crucial hyperparameter:
when setting the threshold on ID (validation) data, this may not distinguish well between ID and OOD samples. 
Conversely, using OOD samples to set this uncertainty threshold \textit{requires access to OOD data}. Not only may 
OOD data not be available, but a threshold set on one OOD dataset may overfit and 
not be suitable for samples from another OOD dataset. Furthermore, even in a situation where several 
sets of OOD data are available, it is unclear whether the uncertainty threshold 
is best set on OOD data closer or further away (in data and/or feature space)
from the ID dataset, or whether and how uncertainty thresholds set on different datasets
should be aggregated. While beyond the scope of this paper, these questions form 
an interesting direction for further research.

\section{Conclusion}
Our work underscores the importance of independent research evaluating state-of-the-art algorithms
on challenging datasets beyond the like of CIFAR-10. 
Based on our work, we expect similar results in other clinical data modalities 
that share characteristics with histopathology, such as high sample variability, small datasets, and 
high levels of label noise.

We demonstrate which uncertainty evaluation methods
can work well on histopathological data, we caution against relying on 
individual calibration metrics (see also Appendix). 
We found that SNGP -- a distance-aware method that is sensitive to 
differences in ID and OOD data -- is the best performing uncertainty estimation method 
on histopathology data.
However, we show that the use of uncertainty thresholding for OOD detection or to increase 
predictive accuracy is problematic in histopathology.

Promising directions for future work will be, first, to test 
other OOD detection methods, and second, to develop clinically meaningful
downstream tasks and datasets on which models and methods can be evaluated.
We hope that
our work helps to bridge the gap between current developments in ML and histopathology, and 
encourage developers of novel uncertainty estimation and OOD detection methods to benchmark these not just on 
standard ML datasets, but also on histopathological data, where their performance is highly 
relevant for clinical applications.  

\section*{Acknowledgements}
We thank Stefan Bauer, Robert Vandersluis, Rafael Poyiadzi, Emma Slade and two anonymous reviewers for helpful discussions and comments.

\newpage
\bibliographystyle{plainnat}
\bibliography{biblio.bib}

\newpage
\section{Appendix}
\subsection{Selective classification results all thresholds -- accuracy}

\begin{table}[h]
  \centering
    \begin{tabular}{|l|l|l|l|l|}
    \hline
    \textbf{threshold set on \textrightarrow}  & \textbf{validation}    & \textbf{hospital 4}  & \textbf{hospital 5}  & \textbf{cancer} \\
    \hline
    \multicolumn{5}{|c|}{Softmax}                   \\
    \hline    
    \textbf{validation}   &   0.984 $\pm$ 0.015 & 1.000 $\pm$ 0.000 & 0.999 $\pm$ 0.001 & 0.999 $\pm$ 0.001  \\
    \textbf{hospital 4}  & 0.765 $\pm$ 0.079 & 0.821 $\pm$ 0.087 & 0.822 $\pm$ 0.086 & 0.821 $\pm$ 0.085 \\
    \textbf{hospital 5}  & 0.624 $\pm$ 0.054 & 0.714 $\pm$ 0.096 & 0.701 $\pm$ 0.115 & 0.714 $\pm$ 0.095 \\
    \textbf{cancer}      & 0.591 $\pm$ 0.130 & 0.668 $\pm$ 0.148 & 0.685 $\pm$ 0.156 & 0.684 $\pm$ 0.134 \\
    \hline
    \multicolumn{5}{|c|}{Dropout}                   \\
    \hline
    \textbf{validation}  & 0.990 $\pm$ 0.001 & 1.000 $\pm$ 0.001 & 1.000 $\pm$ 0.000 & 0.999 $\pm$ 0.001 \\
    \textbf{hospital 4}  & 0.801 $\pm$ 0.052 & 0.932 $\pm$ 0.050 & 0.882 $\pm$ 0.064 & 0.909 $\pm$ 0.040 \\
    \textbf{hospital 5} & 0.633 $\pm$ 0.074 & 0.768 $\pm$ 0.117 & 0.652 $\pm$ 0.150 & 0.748 $\pm$ 0.118 \\
    \textbf{cancer}     & 0.675 $\pm$ 0.102 & 0.643 $\pm$ 0.141* & 0.596 $\pm$ 0.121 & 0.613 $\pm$ 0.135 \\
    \hline
    \multicolumn{5}{|c|}{Ensemble}                   \\
    \hline
    \textbf{validation}   & 0.995 $\pm$ 0.000 & 1.000 $\pm$ 0.000 & 1.000 $\pm$ 0.000 & 1.000 $\pm$ 0.000 \\
    \textbf{hospital 4}   & 0.832 $\pm$ 0.019 & 0.924 $\pm$ 0.039 & 0.955 $\pm$ 0.035 & 0.890 $\pm$ 0.021 \\
    \textbf{hospital 5}  & 0.619 $\pm$ 0.020 & 0.689 $\pm$ 0.028 & 0.717 $\pm$ 0.054 & 0.682 $\pm$ 0.038 \\
    \textbf{cancer}     & 0.637 $\pm$ 0.021 & 0.745 $\pm$ 0.074 & 0.673 $\pm$ 0.090 & 0.787 $\pm$ 0.007 \\
    \hline    
    \multicolumn{5}{|c|}{SNGP}                   \\
    \hline    
    \textbf{validation}  & 0.990 $\pm$ 0.001 & 1.000 $\pm$ 0.000 & 1.000 $\pm$ 0.000 & 1.000 $\pm$ 0.000\\
    \textbf{hospital 4}  & 0.793 $\pm$ 0.065 & 0.961 $\pm$ 0.033 & 0.959 $\pm$ 0.030 & 0.966 $\pm$ 0.034 \\
    \textbf{hospital 55}  & 0.737 $\pm$ 0.038 & 0.870 $\pm$ 0.090 & 0.863 $\pm$ 0.085 & 0.883 $\pm$ 0.098 \\
    \textbf{cancer}     & 0.631 $\pm$ 0.057 & 0.669 $\pm$ 0.081 & 0.655 $\pm$ 0.090 & 0.695 $\pm$ 0.071 \\
    \hline
  \end{tabular}
  \caption{\label{theshold-all-acc-table} Accuracy across all datasets (columns) for uncertainty thresholds set on all 
  datasets (rows), mean $\pm$ std,  averaged over 4 seeds. *averaged over 3 seeds, all samples rejected for 1 seed.}
\end{table}

\subsection{Selective classification results all thresholds -- average precision}

\begin{table}[h]
  \centering
    \begin{tabular}{|l|l|l|l|l|}
    \hline
    \textbf{threshold set on \textrightarrow}  & \textbf{validation}    & \textbf{hospital 4}  & \textbf{hospital 5}  & \textbf{cancer} \\
    \hline
    \multicolumn{5}{|c|}{Softmax}                   \\
    \hline    
    \textbf{validation}   & 0.979 $\pm$ 0.015 & 0.999 $\pm$ 0.001 & 0.999 $\pm$ 0.001 & 0.998 $\pm$ 0.001 \\
    \textbf{hospital 4}  & 0.736 $\pm$ 0.066 & 0.842 $\pm$ 0.162 & 0.777 $\pm$ 0.124 & 0.812 $\pm$ 0.151  \\
    \textbf{hospital 5}  & 0.583 $\pm$ 0.039 & 0.656 $\pm$ 0.086 & 0.619 $\pm$ 0.079 & 0.62 $\pm$ 0.087 \\
    \textbf{cancer}      & 0.12 $\pm$ 0.006 & 0.155 $\pm$ 0.052 & 0.136 $\pm$ 0.023 & 0.157 $\pm$ 0.042  \\
    \hline
    \multicolumn{5}{|c|}{Dropout}                   \\
    \hline
    \textbf{validation}  & 0.986 $\pm$ 0.002 & 0.999 $\pm$ 0.001 & 0.999 $\pm$ 0.000 & 0.999 $\pm$ 0.001 \\
    \textbf{hospital 4} & 0.769 $\pm$ 0.040 & 0.941 $\pm$ 0.062 & 0.881 $\pm$ 0.092 & 0.914 $\pm$ 0.061 \\
    \textbf{hospital 5} & 0.597 $\pm$ 0.057 & 0.644 $\pm$ 0.221 & 0.610 $\pm$ 0.156 & 0.672 $\pm$ 0.210  \\
    \textbf{cancer}     & 0.123 $\pm$ 0.002 & 0.260 $\pm$ 0.107* & 0.201 $\pm$ 0.080 & 0.225 $\pm$ 0.106 \\
    \hline
    \multicolumn{5}{|c|}{Ensemble}                   \\
    \hline
    \textbf{validation}   & 0.993 $\pm$ 0.000 & 1.000 $\pm$ 0.000 & 1.000 $\pm$ 0.000 & 0.999 $\pm$ 0.000  \\
    \textbf{hospital 4} & 0.790 $\pm$ 0.016 & 0.911 $\pm$ 0.046 & 0.951 $\pm$ 0.041 & 0.866 $\pm$ 0.018 \\
    \textbf{hospital 5} & 0.579 $\pm$ 0.015 & 0.603 $\pm$ 0.038 & 0.642 $\pm$ 0.071 & 0.579 $\pm$ 0.034 \\
    \textbf{cancer}     & 0.124 $\pm$ 0.000 & 0.231 $\pm$ 0.089 & 0.310 $\pm$ 0.107 & 0.173 $\pm$ 0.011 \\
    \hline    
    \multicolumn{5}{|c|}{SNGP}                   \\
    \hline    
    \textbf{validation}  & 0.986 $\pm$ 0.002 & 1.000 $\pm$ 0.000 & 1.000 $\pm$ 0.000 & 1.000 $\pm$ 0.000 \\
    \textbf{hospital 4}  & 0.765 $\pm$ 0.055 & 0.870 $\pm$ 0.121 & 0.869 $\pm$ 0.110 & 0.881 $\pm$ 0.118 \\
    \textbf{hospital 5}  & 0.693 $\pm$ 0.032 & 0.613 $\pm$ 0.300 & 0.633 $\pm$ 0.281 & 0.632 $\pm$ 0.316 \\
    \textbf{cancer}     & 0.124 $\pm$ 0.001 & 0.253 $\pm$ 0.063 & 0.262 $\pm$ 0.085 & 0.291 $\pm$ 0.084 \\
    \hline
  \end{tabular}
  \caption{\label{theshold-all-ap-table} Average precision across all datasets (columns) for uncertainty thresholds set on all 
  datasets (rows), mean $\pm$ std,  averaged over 4 seeds. *averaged over 3 seeds, all samples rejected for 1 seed.}
\end{table}

\subsection{Limitations of uncertainty estimation metrics}
While metrics can be indicative of performance, 
ideally, uncertainty estimation methods should be evaluated
in a clinical setting, i.e., whether and by how much the quality of their
uncertainty estimates affects diagnosis and therefore improves patient outcomes.
This is difficult to implement in histopathology:
all ML models intended for use in healthcare, 
in particular if algorithmic predictions directly affect patients, 
would have to demostrate their practical benefits for improving diagnostic accuracy and patient outcomes 
through rigorous embedding and evaluating the model in a clinical trial
\footnote{To date, the U.S. Food and Drug Administration (FDA) has approved
  only one ML-based software, \textit{"Paige Prostate"}.
  Its use is limited to \textit{assisting}
  pathologists in their assessment, and the associated clinical study did not evaluate 
  the software's impact on final patient diagnosis.}.
In addition to the logistical \citep{steiner2021closing} and financial requirements,  
there are currently no established best practices and only minimal, non-technical regulatory frameworks for
such a scenario.  

As a result,
there are currently no histopathology models
used in the clinic on which such an analysis could be performed.
Furthermore, there is no data, such as survival data, that we could have used as a
clinically relevant \textit{"downstream"} task \citep{locatello2019challenging}, 
for the Camelyon17 dataset.
Lacking relevant data for a downstream task, we explored whether 
uncertainty estimates could benefit the classification of a model's 
learnt features into four different cancer stages
\footnote{These were: negative (no metastases);
isolated tumour cells (single tumour cells or a cluster of tumour cells, not larger than 
0.2mm or less than 200 cells);
micro-metastasis (larger than 0.2mm and/or containing more than 200 cells, but not larger
than 2mm); and
macro-metastasis (larger than 2mm). For more details see \citep{bandi2018detection}.}. 
We did not find reliable improvements in classification performance
as a consequence of better uncertainty estimates, and the results were
inconsistent across uncertainty estimation methods and stages.
Thus, while beyond the scope of this paper, developing datasets to evaluate uncertainty estimation methods
on clinically meaningful downstream tasks is another promising
direction for future work.

\end{document}